\documentclass[twoside,11pt]{article}
\usepackage{jair, theapa, rawfonts}

\usepackage{booktabs}
\usepackage{multirow}
\usepackage{multicol}
\usepackage[inline]{enumitem}
\usepackage{amssymb}
\usepackage{mathtools}
\usepackage{subcaption}
\usepackage{xspace}
\usepackage{url}

\usepackage[ruled,linesnumbered,vlined]{algorithm2e}
\newtheorem{definition}{Def}

\usepackage{pgfplots}
\pgfplotsset{compat=1.12}
\usepgfplotslibrary{statistics}
\usepackage{tikz}

\newcommand\VO{\mathit{VO}}
\newcommand\tb{\mathit{TB}}
\newcommand\mc{\mathit{MC}}
\newcommand\mo{\mathit{MO}}
\newcommand{\Rcomb}{$R_{comb}$\xspace}

\newcommand{\rebuttal}[1]{\textcolor{black}{#1}}

\jairheading{82}{2025}{819-850}{06/2023}{02/2025}
\ShortHeadings{Value Preferences Estimation and Disambiguation}
{Liscio, Siebert, Jonker \& Murukannaiah}
\firstpageno{819}

\begin{document}

\title{Value Preferences Estimation and Disambiguation\\ in Hybrid Participatory Systems}

\author{\name Enrico Liscio \email e.liscio@tudelft.nl \\
        \name Luciano C. Siebert \email l.cavalcantesiebert@tudelft.nl \\
       \addr Delft University of Technology, the Netherlands
       \AND
       \name Catholijn M. Jonker \email c.m.jonker@tudelft.nl \\       \addr Delft University of Technology, the Netherlands \and Leiden University, The Netherlands
        \AND
        \name Pradeep K. Murukannaiah \email p.k.murukannaiah@tudelft.nl \\
       \addr Delft University of Technology, the Netherlands
}

\maketitle

\begin{abstract}
Understanding citizens' values in participatory systems is crucial for citizen-centric policy-making. We envision a hybrid participatory system where participants make choices and provide motivations for those choices, and AI agents estimate their value preferences by interacting with them. We focus on situations where a conflict is detected between participants' choices and motivations, and propose methods for estimating value preferences while addressing detected inconsistencies by interacting with the participants. We operationalize the philosophical stance that ``valuing is deliberatively consequential." That is, if a participant's choice is based on a deliberation of value preferences, the value preferences can be observed in the motivation the participant provides for the choice. Thus, we propose and compare value preferences estimation methods that prioritize the values estimated from motivations over the values estimated from choices alone. Then, we introduce a disambiguation strategy that \rebuttal{combines Natural Language Processing and Active Learning to} address the detected inconsistencies between choices and motivations. We evaluate the proposed methods on a dataset of a large-scale survey on energy transition. The results show that explicitly addressing inconsistencies between choices and motivations improves the estimation of an individual's value preferences. The disambiguation strategy does not show substantial improvements when compared to similar baselines---however, we discuss how the novelty of the approach can open new research avenues and propose improvements to address the current limitations.
\end{abstract}

\section{Introduction}
\label{sec:introduction}

Values, \rebuttal{spanning concepts such as self-determination and sustainability,} are the standards or criteria that justify one's opinions and actions and are intrinsically linked to goals \shortcite{Schwartz2012AnValues}. Values form an ordered system of priorities and the relative importance one ascribes to values (one's \emph{value preferences}) guides action. 
\rebuttal{Since values define shared goals and are essential for human cooperation, they are deemed critical to developing AI that can integrate beneficially into our society \shortcite{Russell-2015-AIMagazine-RobustBeneficialAI,Gabriel2020}.}
Yet, how individuals ascribe relative priorities among values can vary significantly across people, socio-cultural environments \shortcite{Dignum2017ResponsibleAutonomy}, and decision contexts \shortcite{Hill2009PersonsDomain}.
\rebuttal{Identifying and reasoning about individuals' value preferences has been recognized as the challenge of \emph{value inference} \shortcite{Liscio2023}, encompassing AI and hybrid human-AI methods proposed to identify the values relevant to a decision-making process \shortcite{Wilson2018,Liscio2022axiesJAAMAS} or to detect values in language \shortcite{Kiesel2022,Qiu2022ValueNet:System}.
Such semi-automated approaches offer the chance to infer individuals' values at a large scale.}

\rebuttal{One crucial field that can benefit from large-scale value inference is policy-making.}
Enhancing citizen participation in decision-making processes is high on the European policy agenda \shortcite{dallhammer2018spatial}. Initiatives to foster citizens' political power and engagement have been proposed through the use of digital platforms for participatory decision-making \shortcite{lafont2015deliberation,Mouter2021a} and deliberation \shortcite{friess2015systematic,iandoli2016online,Shortall2022ReasonDeliberation}. 
To this end, eliciting stakeholders' preferences over competing alternatives only provides superficial information on the debate.
Instead, considering stakeholders' values on a decision-making subject is crucial for crafting long-term policies on the subject \shortcite{Miller2016} since values preferences tend to be stable over time \shortcite{Schwartz2012AnValues}.
\rebuttal{For instance, consider a policy-maker drafting subsidy strategies for solar panels; knowing what value trade-offs motivated the citizens (e.g., sustainability vs. economic efficiency) will inform long-term solutions as well as similar decisions in the future.}

\rebuttal{Within the value inference process, \emph{value preferences estimation} refers to the challenge of estimating an individual's preferences over a given set of relevant values}{\footnote{\rebuttal{Value preferences estimation falls within the realm of descriptive ethics \cite{hamalainen2016descriptive}, aimed at discerning the guiding principles of individuals (with the assumption that they will aim to choose actions that align with their preferred values). It is important to distinguish this from normative ethical theories, such as deontology or utilitarianism, which prescribe how rational agents \textit{ought to} behave. These theories focus on moral decision-making principles and mechanisms, rather than individual value preferences.}}
\rebuttal{\shortcite{Liscio2023}}. \rebuttal{Estimating value preferences on an individual level (as opposed to a population level) allows for (1) a detailed understanding of how different individuals prioritize values; (2) interactive approaches for disambiguation on an individual level. To inform the policy-maker on the population's preferences, value preferences can then be later aggregated at the collective level (e.g., \shortciteauthor{lera2024aggregating}~\citeyear{lera2024aggregating}).}

Value preferences estimation has been traditionally performed based on one's \emph{choices} over competing alternatives, e.g., from answers to value surveys \shortcite{Schwartz2012AnValues,Graham2013} \rebuttal{or from one's action in a context \shortcite{Liscio2023}. In other words, estimating value preferences involves identifying or defining, e.g., through expert assessment or bottom-up aggregation, the relationship between an individual's choice and a value, such as whether the choice promotes the value.}
However, estimating one's value preferences \rebuttal{can be challenging due to the intrinsic uncertainty in defining value-choices relationships and the ambiguity that multiple value preferences could possibly explain a choice \shortcite{Mindermann2018}. Estimating value preferences} 
from both one's choices in a context and the verbal \emph{motivations} for supporting these choices provides additional insights that could not be achieved considering only one source of information. \rebuttal{For instance, consider an individual who recently installed solar panels; they may have been motivated by the values of sustainability or economic efficiency, or both, or neither. Seeking verbal motivations for their choices might unveil their value preferences.} 

We envision a semi-automated approach to value preferences estimation, where AI agents, supported by natural language processing (NLP) techniques, interpret the motivations provided by the participants in support of their choices, and combine the information contained in choices and motivations to estimate their value preferences. But what if the information extracted from the choices conflicts with the information extracted from the motivations given in support of those choices?
To target such conflicts, we propose a hybrid intelligence (HI) \shortcite{Akata2020AIntelligence} approach where value preferences are estimated through the combination of artificial and human intelligence. 

\rebuttal{Consider the aforementioned scenario where citizens provide their choices for subsidy strategies for solar panels. Values such as sustainability and economic efficiency are relevant factors that might influence individuals’ decisions to install solar panels in their homes. Let us assume that an individual supports subsidy policy \emph{A}, for which the value of sustainability was deemed relevant (e.g., by looking into previous decisions and motivations from other individuals, or through expert input). This choice alone does not necessarily reveal their underlying reasons. However, if they solely mention economic efficiency when motivating policy \emph{A}, but not sustainability, a conflict arises.}
We target conflicts between choices and motivations through value preferences \emph{estimation} and \emph{disambiguation}, as shown in Figure~\ref{fig:hps_model}.

\begin{figure}[htb]
    \centering
    \includegraphics[width=0.9\linewidth]{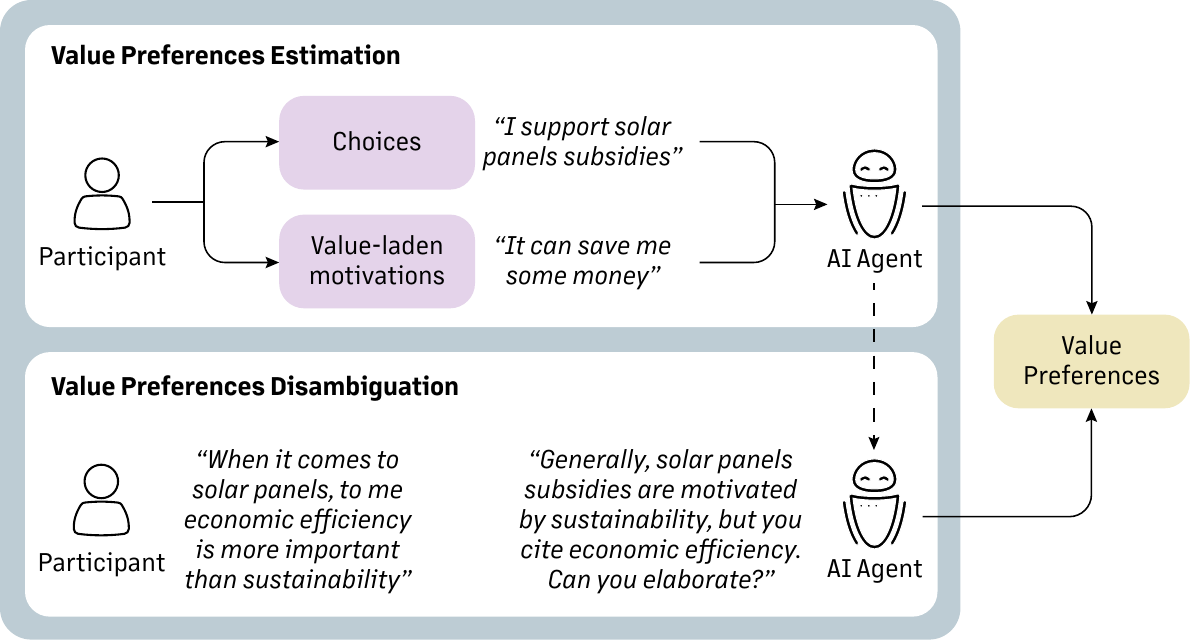}
    \caption{A hybrid participatory system where human participants make choices and motivate those choices, and AI agents estimate \rebuttal{and disambiguate} participants' value preferences.}
    \label{fig:hps_model}
\end{figure}

We propose and compare five methods for estimating value preferences from the choices and motivations provided by participants in a participatory system. \rebuttal{These methods combine participants' choices and motivations in different ways. First, we explore methods that only consider the choices or the motivations. Next, we propose three methods that employ a combination thereof. When choice-motivation conflicts arise,} these methods follow the philosophical account that ``valuing is deliberatively consequential"\rebuttal{\footnote{\rebuttal{In this context, `consequential' refers to one's disposition to treat certain value-related considerations as a reason for action. It should not be confused with 'consequentialism' as a normative ethical theory that asserts the consequences of one's actions should be the primary basis for moral judgments.}} \shortcite{Scheffler2012Valuing}, i.e., if one’s choice is based on a deliberation of value preferences, the value preferences can be observed in the motivation provided for the choice \shortcite{Dietz1995TowardConstruction,kenter2016deliberative,Pigmans2019TheGovernance}.
Thus, the methods prioritize the values observed in the motivations over those observed in the choices.}

\rebuttal{Nevertheless, the detected choice-motivation conflicts ought to be addressed. Such conflicts} may be caused by
\begin{enumerate*}[label=(\arabic*)]
    \item mistakes in the value preferences estimation process (e.g., misclassification of the values supporting the participants' motivations by an NLP model), or
    \item genuine inconsistencies between the participants' choices and motivations, e.g., due to participants having different assumptions regarding values that drive a choice, or due to the value-action gap \cite{franco2022shapes}.
\end{enumerate*}
In both cases, addressing the inconsistencies can be beneficial. If the inconsistency is caused by a mistake in the automatic value preferences estimation process, the involved participant should be asked to resolve the mistake, e.g., by correcting a misclassification of the NLP model. In case the interpretation is confirmed to be correct and the inconsistency is accurately detected, the participant can be guided through a process of self-reflection \shortcite{Lim2019FacilitatingConditions,Liscio2023} and offered the chance to change their choices or provide additional motivations.

\rebuttal{In participatory systems, not all participants may be available to take part in such interactions, and the required additional effort may dissuade participants from engaging \shortcite{Shortall2022ReasonDeliberation}.} 
Inspired by Active Learning (AL) \shortcite{Settles+2012+AL}, we propose a disambiguation strategy that guides the interactions between AI agents and participants, following the rationale that, by addressing the most informative participants first, the quality of value preferences estimation should rapidly improve for all participants. Accordingly, the strategy iteratively selects the participants whose value preferences estimated solely from their choices are most different from the value preferences estimated solely from their motivations. We test this strategy by retrieving the correct interpretation of the motivations provided by the selected participants (i.e., the correct values that support their motivations) to iteratively improve the NLP model tasked to predict the values that support the participants' motivations, which are in turn used to estimate their value preferences.

\paragraph{Contribution}
We propose a method for estimating individuals' value preferences through a disambiguation strategy. Our method is composed of two parts.
First, we propose and compare five methods for estimating individual value preferences. We employ the proposed methods to estimate the value preferences of the participants \rebuttal{in} a large-scale survey on energy transition \shortcite{Spruit2020}, \rebuttal{where participants allocate a number of points among policy options and textually motivate their choices}. We evaluate the extent to which our methods' estimations concur with those of human evaluators. Our results show that addressing the inconsistencies between choices and motivations improves the estimation of value preferences.
Second, we propose and evaluate a disambiguation strategy that is driven by inconsistencies between participants' choices and motivations. We evaluate the strategy in an active learning setting with the value-annotated survey on energy transition, and compare it to traditional NLP AL strategies.
We show that our method leads to comparable results to the tested baselines, both in NLP performance and value preferences estimation. We discuss the results and elaborate on future directions.

\paragraph{Extension} This paper extends the conference paper from Siebert et al. \citeyear{Siebert2022}, where we propose and compare five methods for estimating value preferences from one's choices and motivations. We extend the work by introducing the disambiguation strategy, which naturally complements the value preferences estimation methods in two ways. First, it relies on the same philosophical account by addressing inconsistencies between one's choices and motivations. Second, it situates the value preferences estimation endeavor in a realistic setting, by proposing a concrete and scalable approach for performing value preferences estimation \emph{during} a participatory process. We extend the evaluation in the original paper to validate our proposed disambiguation strategy by using it as a sampling strategy in an AL setting and comparing it to traditional sampling strategies.

\paragraph{Organization} Section~\ref{sec:rel-works} discusses related works. Section~\ref{sec:background} introduces the context behind our analyses. Section~\ref{sec:methods} describes the methods we propose for value preferences estimation and for the disambiguation strategy. Section~\ref{sec:experimental-setting} describes our experimental setup and Section~\ref{sec:results} presents our results. Finally, \rebuttal{Section~\ref{sec:limitations} discusses the limitations of our work} and Section~\ref{sec:conclusions} concludes the paper.

\section{Related Works}
\label{sec:rel-works}

We discuss related works on \rebuttal{valuing,} estimating value preferences, NLP techniques for classifying values from text, and active learning.

\subsection{\rebuttal{Valuing}}
\label{sec:valuing}

\rebuttal{\shortciteauthor{smith1989dispositional}~\citeyear{smith1989dispositional} describe valuing as a form of desiring. However, this conception is limited. For instance, to value someone's leadership skills does not mean desiring this person's leadership skills for oneself, but recognizing them as something positive. Thus, some philosophers have rejected the reduction of valuing to desiring and proposed that it should be perceived in a broader sense, as having a favorable attitude towards something, involving both reason and emotion \shortcite{Scheffler2012Valuing,frankfurt2018freedom}.}

\rebuttal{\shortciteauthor{Scheffler2012Valuing}~\citeyear{Scheffler2012Valuing} suggests that having such a favorable attitude towards something implies a deliberative significance. That is, to value $X$ involves not only seeing $X$ as a source of reasons for action (a view also supported by \shortciteauthor{Schwartz2012AnValues}~\citeyear{Schwartz2012AnValues}) but also having considerations related to $X$ in a relevant context. For example, if one values \textit{privacy}, they would contemplate in relevant contexts to treat considerations about the impact of proposed actions on their privacy as having deliberative importance. In other words, \shortciteauthor{Scheffler2012Valuing} states that valuing is \textit{deliberatively consequential}.}

\rebuttal{Several researchers support this view. For instance, \shortciteauthor{Dietz1995TowardConstruction}~\citeyear{Dietz1995TowardConstruction} consider the notion that people assess their options in terms of expected outcomes (subjective expected utility model), referring to personal values. \shortciteauthor{kenter2016deliberative}~\citeyear{kenter2016deliberative} propose the Deliberative Value Formation model, in which deliberation is considered to form values through processes that may inform and enable reflection. In the context of citizen participation, \shortciteauthor{Pigmans2019TheGovernance}~\citeyear{Pigmans2019TheGovernance} suggest that, if values that stakeholders perceive as relevant can be identified as part of the deliberation process, reflection and mutual understanding could be promoted.}

\rebuttal{In this work, we follow \shortciteauthor{Scheffler2012Valuing}~\citeyear{Scheffler2012Valuing} and others by taking the stance that, if one’s choice is based on a deliberation of value preferences, the value preferences can be observed in the motivation provided for the choice. This approach can support increasing legitimacy in decision-making, by providing a grounded approach for estimating value preferences.\footnote{\rebuttal{In this work, we do not aim to model individual moral decision-making, e.g., as discussed by \shortciteauthor{haidt2001emotional}~\citeyear{haidt2001emotional}, who argues that moral choices are based on intuitions rather than reasoning or deliberation. Instead, we focus on valuing as a deliberative process to support and legitimize participation.}}}

\subsection{Estimating Value Preferences}

Survey instruments such as the Portrait Value Questionnaire \shortcite{Schwartz2012AnValues}, Schwartz Value Survey \shortcite{Schwartz2012AnValues}, Value Living Questionnaire \shortcite{Wilson2010TheFramework}, and Moral Foundations Questionnaire \shortcite{graham2011mapping} have been used to estimate an individual's preferences towards a set of values.
Further, some approaches combine self-reported surveys with participatory design \shortcite{Pommeranz2011,Liao2019EnablingFictions}, following the principles of Value Sensitive Design \shortcite{Friedman2019ValueImagination}.
However, value questionnaires have been criticized for being incomplete and not context-sensitive \shortcite{LeDantec2009,Boyd2015ValuesValues}. In this work, we do not query participants directly about their value preferences, but evaluate their choices and related motivations in context.

Alternatively, value preferences can be estimated from a bottom-up approach by analyzing human behavior and choices. In the field of economics, values have been elicited via revealed preference methods such as direct elicitation and multiple price lists \shortcite{Benabou2020ElicitingExperiment}. For complex and high-dimensional environments, inverse reinforcement learning algorithms \cite{Ng2000}, which focus on extracting a ``reward function'' given observed optimal behavior, show promising results \shortcite{Russell2019HumanControl}. However, critiques on the infeasibility of estimating an individual's rationality and preferences (including value preferences) simultaneously \shortcite{Mindermann2018} suggest the need for additional normative assumptions\rebuttal{, e.g., an explicit model of the cognitive processes that guided a given behavior}. \rebuttal{Furthermore, the use of reward or objective functions has been argued not to be well-suited for modeling human values or other normative concepts (i.e., judgments of what is right, wrong, good, or bad) \shortcite{eckersley2018impossibility}.} 
We seek to address such critiques by \rebuttal{(1)} incorporating textual motivations provided by humans for their choices and using NLP approaches to automatically classify the values that underlie the motivations \rebuttal{and (2) using partially ordered preferences for modeling value preferences}.

\subsection{Classifying Values from Text}

A classical approach to value classification from text is through value dictionaries---lists of word characteristic of certain values---by measuring the relative frequency of the words describing each value \shortcite{Pennebaker2001} e.g., the Moral Foundation Dictionary \shortcite{Graham2013}. These dictionaries have been expanded through semi-automated methods \shortcite{Wilson2018,Araque2020,Hopp2020} or through NLP techniques \shortcite{Ponizovskiy2020,Araque2021}, and limitations related to word count techniques have been approached via word embedding models \shortcite{Garten2018,Bahngat2020,Pavan2020MoralityText}. More recent approaches use supervised machine learning \shortcite{Liscio2022a,Kiesel2022,Alshomary2022,Huang2022,Liscio2023ACL,van-der-meer-etal-2023-differences,park-etal-2024-morality}, where NLP models are trained on datasets annotated with value taxonomies, such as the Moral Foundation Twitter Corpus \shortcite{Hoover2020} and ValueNet \shortcite{Qiu2022ValueNet:System}.
Our method builds on this approach, as we train an NLP model on an annotated dataset. However, we expand on the literature by employing an active learning approach.

\subsection{Active Learning}
\label{sec:active-learning-related-works}

The key idea behind Active Learning (AL) is that a supervised ML algorithm can achieve good performance with few training examples if such examples are suitably selected \shortcite{Settles+2012+AL}.
In a traditional AL setting, a large set of \emph{unlabeled data} is available, and an \emph{oracle} (e.g., human annotators) can be consulted to annotate the unlabeled data. A \emph{sampling strategy} is used to iteratively select the next batch of unlabeled data to be annotated by the oracle, with the intent of rapidly improving the performance of the ML algorithm.
A commonly used sampling strategy is uncertainty sampling \shortcite{Ren2021}, where at every iteration the ML algorithm is used to predict labels on all the unlabeled data, and the $m$ unlabeled data with the highest label entropy are selected as the next batch to be annotated (i.e., the data on which the model is least confident about its prediction).

AL has been extensively used in NLP applications \shortcite{Zhang2022}, with two main strategy approaches. On the one hand, some strategies use the \emph{informativeness} of each unlabeled instance individually, e.g., by measuring the uncertainty of the prediction or the norm of the gradient \shortcite{Zhang2017a}. The unlabeled instances that are estimated to be most informative are selected to be labeled by the oracle.
On the other hand, other strategies focus on the \emph{representativeness} of the data, e.g., by selecting data points that are most representative of the unlabeled set \shortcite{Zhao2020} or that are most different from the data that is already labeled \shortcite{Erdmann2019}.
In general, state-of-the-art AL strategies exploit information about the NLP task (i.e., about the NLP model and the available data) with the intent of rapidly improving the performance of the NLP model. However, in our setting, the NLP model is a means to the end of estimating value preferences. Hence, we propose a strategy that is driven by the informativeness of the unlabeled data, but where the informativeness is derived by the downstream task of value preferences estimation.

\section{Background}
\label{sec:background}

We introduce the dataset and formalize the key concepts to provide a background for our methods and experiments.

\subsection{Participatory Value Evaluation (PVE)}
\label{sec:pve}

We estimate individual value preferences from choices and motivations provided via Participatory Value Evaluation (PVE) \shortcite{Mouter2021a}, an online participatory system.
We use data from a PVE conducted between April and May 2020 involving 1376 participants \shortcite{Spruit2020}, aimed at supporting the municipality of Súdwest-Fryslân in the Netherlands in co-creating an energy transition policy, increasing citizen participation, and avoiding public resistance as happened in previous projects on sustainable energy \shortcite{DutchMinistryofEconomicAffairsandClimate2019}. The main question to the citizens was: ``What do you find important for future decisions on energy policy?" Six policy options (Table \ref{tab:choice-options}) were developed in consultation with 45 citizens. These options were presented in the PVE platform, with the participants asked to distribute 100 points among them. 
\rebuttal{In most cases, participants assigned points to more than one option, with options $o_1$ and $o_2$ receiving more than half of the points on average.}
After dividing the points, the participants had the chance to \rebuttal{provide a textual motivation in support of each of the options to which they had allocated points}. 876 participants provided at least one motivation for their choices, resulting in a total of 3229 motivations.

\begin{table}[htb]
\small
\centering
\begin{tabular}{@{}c@{\hspace{.2cm}}p{8.4cm}@{}c@{}p{0.8cm}@{}}
\toprule
\textbf{Policy option} & \textbf{Description} & \textbf{Avg. points distributed} \\ 
\midrule
$o_1$ & The municipality takes the lead and unburdens you & 29.05      \\
$o_2$ & Inhabitants do it themselves & 21.72      \\
$o_3$ & The market determines what is coming & 9.39 \\
$o_4$ & Large-scale energy generation will occur in a small number of places & 15.01 \\
$o_5$ & Betting on storage (Súdwest-Fryslân becomes the battery of the Netherlands) & 12.96 \\
$o_6$ & Become a major energy supplier in the Netherlands & 4.71  \\
\bottomrule
\end{tabular}
\caption{Policy options available in the energy transition PVE.}
\label{tab:choice-options}
\end{table}

\rebuttal{The motivations were annotated with the underlying values as part of the original data collection. We refer to \shortciteauthor{kaptein2020participatory}~\citeyear{kaptein2020participatory} for a detailed description of the annotation procedure, which we summarize here.}
The values embedded in the textual motivations were identified by a team of four annotators using a grounded theory approach \shortcite{Heath2004DevelopingStrauss}. The annotators were first introduced to foundational concepts \shortcite{Schwartz2012AnValues,Graham2013} and examples of values. Then, they were asked to annotate any keywords from the motivations that relate to values. After a consolidation round, annotators agreed on a list with 18 values \rebuttal{(as presented in Appendix A.1)}. In this paper, we consider only the most frequent values (values mentioned at least 250 times across all project options) to demonstrate our methods. \rebuttal{This allows us to perform an in-depth analysis and provide a digestible overview of the results while managing the computational load. Nevertheless, our methods are agnostic of the number of values, as we further discuss in Section~\ref{sec:limitations}.} Table~\ref{tab:selected-values} shows the value list we consider in our experiments.

\begin{table}[htb]
\centering
\small
\begin{tabular}{@{}c@{\hspace{.2cm}}p{2.3cm}@{\hspace{.2cm}}p{10.9cm}@{}}
\toprule
\textbf{Value ID} & \textbf{Value name} & \textbf{Description}  \\ 
\midrule
$v_1$ & Cost-effectiveness & Money must be well spent and the project must be profitable. No waste. Costs should not be too high   \\
$v_2$ & Nature and\newline landscape &  Nature and environment are important. Horizon pollution is often seen as negative. Preserving the Frisian landscape is central      \\
$v_3$ & Leadership & Clarity and control over the sustainability of the energy system. Often about an organization or person that has to take charge \\
$v_4$ & Cooperation &   Working together on a goal. Residents can work together, but also groups and organizations\\
$v_5$ & Self-determination & The opportunity for residents to make their own decision on renewable energy and to be able to implement it\\
\bottomrule
\end{tabular}
\caption{Considered values for the energy transition PVE.}
\label{tab:selected-values}
\end{table}

Table~\ref{tab:annotation} shows the number of annotations provided for each of the values we analyze (described in Table~\ref{tab:selected-values}). Although all values have more than 250 annotations (our selection criterion), these values were not annotated equally across the choice options. For example, $v_3$ was annotated 349 ($\sim$76\%) times for $o_3$, and only 3 times for $o_6$.

\begin{table}[htb]
\centering
\small
\begin{tabular}{lllllllll}
\centering
& & \multicolumn{6}{c}{\textbf{Options}} \\ 
\multirow{5}{*}{{\rotatebox[origin=b]{90}{\textbf{Annotated values}}}} & & $o_1$ & $o_2$ & $o_3$ & $o_4$ & $o_5$ & $o_6$ & \textbf{$O$} \\ \cline{3-9} 
& \multicolumn{1}{l|}{$v_1$} & \multicolumn{1}{l|}{90} & \multicolumn{1}{l|}{85} & \multicolumn{1}{l|}{102} & \multicolumn{1}{l|}{85} & \multicolumn{1}{l|}{89} & \multicolumn{1}{l|}{58} & \multicolumn{1}{l|}{\textbf{509}} \\ \cline{3-9} 

& \multicolumn{1}{l|}{$v_2$} & \multicolumn{1}{l|}{50} & \multicolumn{1}{l|}{29} & \multicolumn{1}{l|}{11} & \multicolumn{1}{l|}{269} & \multicolumn{1}{l|}{27} & \multicolumn{1}{l|}{47} & \multicolumn{1}{l|}{\textbf{433}}\\ \cline{3-9} 

& \multicolumn{1}{l|}{$v_3$} & \multicolumn{1}{l|}{349} & \multicolumn{1}{l|}{40} & \multicolumn{1}{l|}{42} & \multicolumn{1}{l|}{13} & \multicolumn{1}{l|}{11} & \multicolumn{1}{l|}{3} & \multicolumn{1}{l|}{\textbf{458}}\\ \cline{3-9} 

& \multicolumn{1}{l|}{$v_4$} & \multicolumn{1}{l|}{80} & \multicolumn{1}{l|}{131} & \multicolumn{1}{l|}{35} & \multicolumn{1}{l|}{17} & \multicolumn{1}{l|}{13} & \multicolumn{1}{l|}{31} & \multicolumn{1}{l|}{\textbf{307}}\\ \cline{3-9} 

& \multicolumn{1}{l|}{$v_5$} & \multicolumn{1}{l|}{35} & \multicolumn{1}{l|}{305} & \multicolumn{1}{l|}{7} & \multicolumn{1}{l|}{8} & \multicolumn{1}{l|}{20} & \multicolumn{1}{l|}{16} & \multicolumn{1}{l|}{\textbf{391}}\\ \cline{3-9} 

& \multicolumn{1}{l|}{\textbf{$V$}} & \multicolumn{1}{l|}{\textbf{604}} & \multicolumn{1}{l|}{\textbf{590}} & \multicolumn{1}{l|}{\textbf{197}} & \multicolumn{1}{l|}{\textbf{392}} & \multicolumn{1}{l|}{\textbf{160}} & \multicolumn{1}{l|}{\textbf{\textbf{155}}} \\ \cline{3-8} 
\end{tabular}
\caption{Distribution of values annotated for each policy option.}
\label{tab:annotation}
\end{table}

\subsection{Formalization}

We formalize the concepts associated with the PVE (choices and motivations) and with value preferences estimation (value systems and value-option matrix). These concepts are related as shown in Figure~\ref{fig:ch-mot-val}.

\begin{figure}[htb]
    \centering
    \small
    \includegraphics[width=\textwidth]{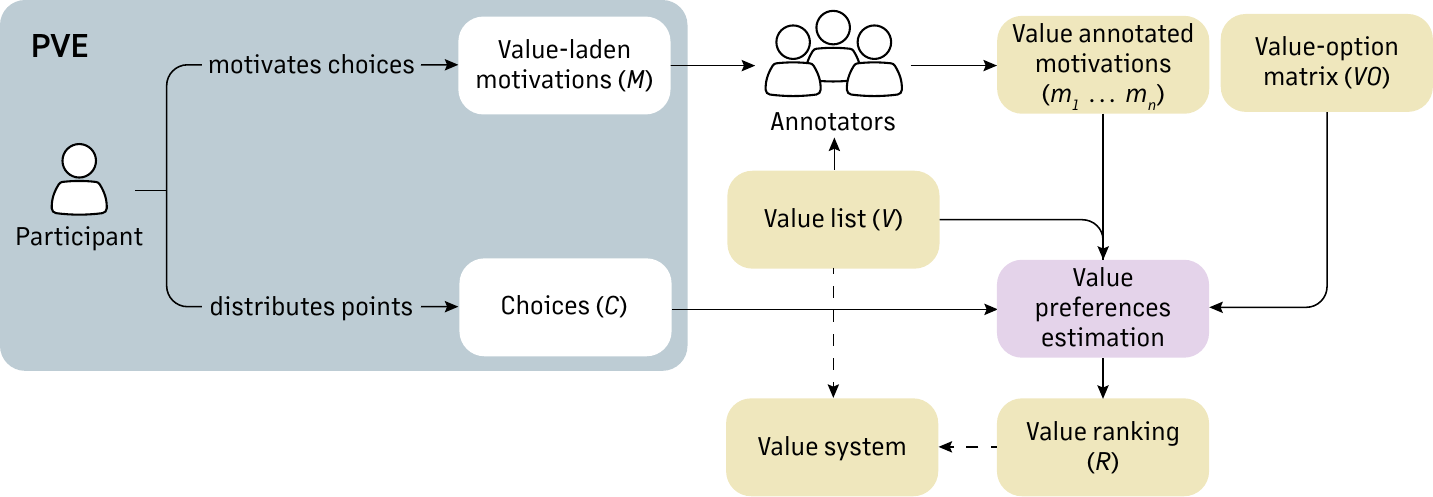}
    \caption{\rebuttal{Each PVE participant makes choices $C$ (i.e., distributes points to the policy options) and provides motivations $M$ to their choices. The participant's value system is defined as the ranking $R$ over a set of values $V$. Our proposed value preferences estimation methods estimate $R$ based on (1) a given value list $V$, (2) the choices $C$, (3) the values annotated in the motivations, and (4) an initial estimate of their value-option matrix $VO$.}}
    \label{fig:ch-mot-val}
\end{figure}

\subsubsection{Value System}
\label{sec:value-system}

Values can be ordered according to their subjective importance as guiding principles \shortcite{Schwartz2012AnValues}. Each person has a \textit{value system} that internally defines the importance the values have to a person according to their preference and context. We represent this value preference via a ranking \shortcite{Zintgraf2018OrderedMaking}. Adapting from \shortciteauthor{Serramia2021}~\citeyear{Serramia2021}, we formally define a value system as follows.

\begin{definition}
A value system is a pair $\langle V, R\rangle$, where $V$ is a non-empty set of values, and $R$ is the ranking of $V$ which represents a person's value preference.
\end{definition}

\begin{definition} 
\label{def:ranking}
A ranking $R$ of $V$ is a reflexive, transitive, and total binary relation, noted as $v_a \succeq v_b$. Given $v_a,v_b \in V$, if $v_a \succeq v_b$, we say $v_a$ is more preferred than $v_b$. If $v_a \succeq v_b$ and $v_b \succeq v_a$, then we note it as $v_a \sim v_b$ and consider $v_a$ and $v_b$ indifferently preferred. However, if $v_a \succeq v_b$ but it is not true that $v_b \succeq v_a$ (i.e., $v_a \neq v_b$), then we note it as $v_a \succ v_b$.
\end{definition}

In this work, we fix the set of values $V$ for all participants (see Table~\ref{tab:selected-values}) and we propose methods to estimate individuals' rankings over $V$. We refer to this task as \emph{value preferences estimation} in the remainder of the paper.
Further, ranking as defined here allows us to know the preferences between any pair of elements (unlike partial orders). We recognize that one's value preferences might not be a total order, since one could consider a given set of values incomparable. Yet, we focus on total orders as an initial step in estimating value preferences, given the challenges of fairly aggregating partial orders \shortcite{pini2005aggregating}.

\subsubsection{Choices and Motivations}
\label{subsec:choices_motivations}

Our goal is to estimate an individual $i$'s value preferences via a ranking, $R^i$, from $i$'s choices and the motivations provided for these choices. Let $O = \{o_1,\rebuttal{\dots o_j,}\dots o_n\}$ be a set of $n$ options that $i$ can choose from in a specific context (for example, the policy options presented in Table~\ref{tab:choice-options}). We assume that $i$ indicates their preferences, $C^i$, among the choices in $O$ by distributing a certain number of points, $p$, among the options in $O$.
\[
	C^i = \{ c_1, \rebuttal{\dots c_j,} \dots c_n \}, \quad \rebuttal{c_j} \in [0, p], \quad \sum \rebuttal{c_j} = p 
\]

Let $M^i$ be the set of motivations that $i$ provides for their choices:
\[
    M^i = \{ m_1, \rebuttal{\dots m_j,} \dots m_n \}
\]

Following the premise that valuing is deliberatively consequential, if an individual's value system influences their choice $c_j$, we expect them to mention the values that support choice $c_j$ in the motivation provided. Thus, we represent a motivation $m_j$ as the set of values (for example, \rebuttal{a subset of} the values in Table \ref{tab:selected-values}) that are mentioned in the motivation \rebuttal{(with the set being empty if $i$ assigned no points to $o_j$ (i.e., $c_j=0$) and thus no motivation was provided for that policy option)}:
\[
    m_j = \{ v_1, \rebuttal{\dots v_l,} \dots v_m\}, ~\text{if}~ \rebuttal{v_l \in V} ~\text{influenced}~ \rebuttal{c_j},\rebuttal{~\text{with}~m_j = \varnothing~\text{if}~ c_j = 0}
\]

\subsubsection{Value-Option Matrix}
\label{sec:VO-matrix}

\rebuttal{We define a value-option matrix as follows:}

\begin{definition}
An individual's $i$ value-option matrix $\VO^i$ is a binary matrix with $|V|$ (number of values) rows and $|O|$ (number of options) columns, where:
\[
    \VO^i(v,o)= 
\begin{dcases}
    1,& \text{if value } v \text{ is relevant for option } o\\
    0,& \text{otherwise.}
\end{dcases}
\]
\end{definition}

\rebuttal{We employ $\VO^i$ as a fine-grained representation of an individual $i$'s value preferences that displays which values are relevant for which option for that individual. In the following section, we describe how the proposed value preferences estimation methods employ and adjust $\VO^i$ to compute the individual's value ranking $R^i$.}

\section{Method}
\label{sec:methods}

We propose a method for estimating value preferences through a disambiguation strategy. In this section, we present the two components of our method: the estimation of value preferences and the disambiguation strategy.

\subsection{Value Preferences Estimation}
\label{sec:estimating-value-system}

Our goal is to estimate an individual's $i$ value ranking $R^i$ from the division of points across a set of \emph{choices} and the textual \emph{motivations} provided to each choice.

\rebuttal{First, we propose two methods that compute $R^i$ based either on $i$'s motivations (method $M$, resulting in $R^i_M$) or on $i$'s choices (method $C$, resulting in $R^i_C$). We employ these two methods as baselines. 
Next, we propose three methods that combine choices and motivations. Method $\tb$ (resulting in $R^i_\tb$) resolves ties in $R^i_C$ by using the motivations provided by $i$. Method $\mc$ (resulting in $R^i_\mc$) and method $\mo$ (resulting in $R^i_\mo$) update $VO^i$ by addressing the inconsistencies between choices and motivations and between motivations provided for different policy options, respectively.
Figure~\ref{fig:overview} shows the main elements of the five methods, which are described in detail in the remainder of this subsection.
These methods can be applied sequentially---however, the order in which they are applied can change the final ranking.
Furthermore, methods $C$, $\tb$, $\mc$, and $\mo$ take as input an initial estimate of $\VO^i$ (which gets updated in the case of $\mc$ and $\mo$). We elaborate on the choice of the initial $\VO^i$ in our experiments in Section~\ref{sec:eval-value-preferences}.}

\begin{figure}[ht]
    \centering
    \includegraphics[width=0.95\textwidth]{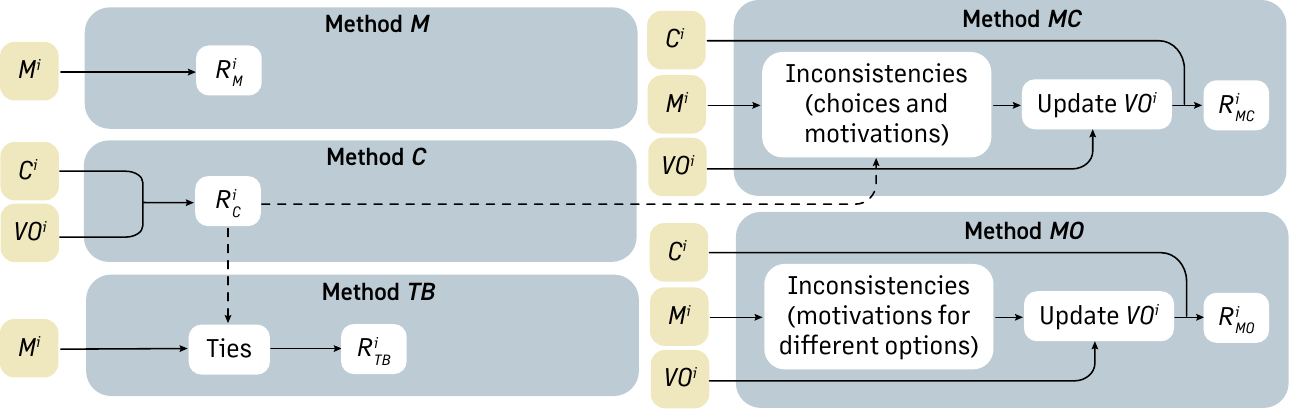}
    \caption{Overview of the five proposed value preferences estimation methods.}
    \label{fig:overview}
\end{figure}

\subsubsection{Method $M$}
\label{sec:method-M}

To estimate an individual's value ranking $R^i_M$ solely based on the motivations $M^i$ provided to their choices $C^i$, we first count how many times a given value is mentioned (i.e., annotated) in any of the motivations provided, and attribute one point to each time it is mentioned. Then, we infer the ranking $R^i_M$ by ordering the values accordingly.

\subsubsection{Method $C$}
\label{sec:method-C}

To estimate an individual's value ranking $R^i_C$ solely based on their choices $C^i$ (vector of size $|O|$, i.e., the number of options), we assume that the individual's choices completely align with their value preferences. First, we compute the importance of the values ($U^i$) for the individual by weighing the values supported by each option ($o_j$) with the points ($c_j$) the individual assigns to the option. Then, we infer a ranking $R^i_C$ from $U^i$, by ordering the values in $V$ according to their importance score in $U^i$.

\begin{align}
    U^i &= \VO^i \times C^{i^T} \label{eq:utility} \\
    R^i_C &= \texttt{rank} (U^i) \label{eq:Rank}
\end{align}

\noindent \rebuttal{where $U^i$ is a $V$-sized vector of non-negative integers. For instance, assume that $i$ distributes their points to the six policy options as $C^i=\{10,20,30,20,0,20\}$ and the initial estimate of $\VO^i$ is as shown in Table~\ref{tab:value-option-matrix}. The multiplication of $VO^i \times C^{i^T}$ results in $U^i=\{100,70,60,80,30\}$.
$R^i_C$ is determined by ordering the values in $V$ according to their importance score: $v_1 \succ v_4 \succ v_2 \succ v_3 \succ v_5$.}

\subsubsection{Method $TB$: Motivations as Tie Breakers}

We use the motivations $M^i$ as \textit{tie breakers} to reduce indifferent preferences in a value ranking. We start with a given ranking $R^i$ (e.g., $R^i_C$). Then, let us define that a tie $\tau_{a,b} \in R^i$ between two values $v_a,v_b \in V$ is present when $v_a$ and $v_b$ are indifferently preferred ($v_a \sim v_b$). 
If there is a tie $\tau_{a,b}$ and if at least one of the motivations mentions $v_a$ but none of the motivations mention $v_b$, then the $TB$ method considers $v_a \succ v_b$, and thus breaks the tie. If both values are mentioned in one of the motivations or not mentioned in any motivation, the tie remains. Algorithm~\ref{alg:tie-breaker} illustrates this method.

\begin{algorithm}[!htb]
\small
\SetAlgoLined
\KwIn{$R^i$, $M^i$}
\KwOut{$R^i_\tb$}
$R^i_\tb \leftarrow R^i$\\
\For{$\tau_{a,b} \in R^i$}{
    \uIf{$\left( \exists m \in M^i: \; v_a \in m \right) \ \land \ \left(\nexists m \in M^i: \; v_b \in m\right)$}{
        set $v_a \succ v_b$ in $R^i_\tb$ \;
    }
    \uElseIf{$\left(\exists m \in M^i: \; v_b \in m\right) \ \land \ \left(\nexists m \in M^i: \; v_a \in m\right)$}{
        set $v_b \succ v_a$ in $R^i_\tb$ \;    
    }}
\caption{Method $TB$}
\label{alg:tie-breaker}
\end{algorithm}

\rebuttal{For instance, assume that $i$ distributes their points to the six policy options as $C^i=\{30,40,10,20,0,0\}$. The multiplication of $VO^i \times C^{i^T}$ returns $U^i=\{100,90,80,80,70\}$, resulting in $R^i_C : v_1 \succ v_2 \succ v_3 \sim v_4 \succ v_5$. However, if one of the motivations provided by the participant mentions $v_4$ and no motivations mention $v_3$, then the $TB$ method breaks the tie by setting $v_4 \succ v_3$, thus resulting in $R^i_{\tb}: v_1 \succ v_2 \succ v_4 \succ v_3 \succ v_5$.}

\subsubsection{Method $MC$: Motivations are More Relevant than Choices}
\label{sec:MC}

There may be an inconsistency between $R^i$ previously estimated for an individual and the values supported by their motivations. That is, $R^i$ indicates $v_b \succ v_a$ but $v_a$ is supported in a motivation $m_j \in M^i$, and $v_b$ is not supported in any motivation. In this case, the $MC$ method prioritizes the value mentioned in the motivation over the one not mentioned, assuming that the value not mentioned is not relevant for individual $i$ in option $o_j$.

When an inconsistency is detected, we assume that the initial value-option matrix $\VO^i$ was inaccurate and update it. In particular, we set the cell of $\VO^i$ corresponding to $v_b$ for the option $o_j$ supported by $m_j = \{v_a\}$ to $0$. 
\rebuttal{For instance, assume that a participant allocates points to option $o_5$ where, according to the initial estimate of $\VO^i$ (as presented in Table~\ref{tab:value-option-matrix}), $v_1$ is relevant but $v_4$ is not, but mentions $v_4$ in the motivation. Then, $\mc$ adjusts $\VO^i$ by setting the cell $(v_1,o_5)$ to 0. We repeat this process for all $v_b:v_b \succ v_a$.}
Once $\VO^i$ is updated for all inconsistencies, we compute the value ranking $R_{MC}^i$ as Algorithm~\ref{alg:MC} illustrates.

\begin{algorithm}[!htb]
\small
\SetAlgoLined
\KwIn{$R^i$, $M^i$, $\VO^i$, $V$, $C^i$}
\KwOut{$R^i_{\mc}$}
\For{$m_o \in M^i$}{
    \For{$v_a \in m_o$}{
        \For{$v_b \in V \setminus \{ v_a \}$}{
            \uIf{$v_a \prec v_b$}{
                $\VO^i(v_b, o) = 0$;
            }
        }
    }
}
$U^i = \VO^i \times C^{i^T}$\;
$R^i_{\mc} \, = \texttt{rank} (U^i)$\;
\caption{Method $MC$}
\label{alg:MC}
\end{algorithm}

\subsubsection{Method $MO$: Motivations are Only Relevant for One \rebuttal{Policy} Option}
\label{sec:MO}

The motivations $M^i$ provided for different \rebuttal{policy} options can also bring inconsistencies. 
\rebuttal{For example, consider the initial estimate of $\VO^i$ as in Table~\ref{tab:value-option-matrix}.}
Further, assume that individual $i$ motivated $o_1$ with value $v_3$ ($m_{1} = \{v_3\}$), and $o_2$ with value $v_5$ ($m_{2} = \{v_5\}$). From the notion of valuing as a deliberatively consequential process, from $m_1$ we can infer that $v_3 \succ v_5$, whereas from $m_2$ we can infer that $v_5 \succ v_3$. 

As in the $MC$ method, when an inconsistency is detected, we assume that the initial value-option matrix $\VO^i$ was inaccurate and update it. In particular, we set the cell of $\VO^i$ corresponding to the value which is part of the inconsistency but was not mentioned in the provided motivation to 0. From our example, the method would set $\VO^i(v_5,o_1)$ and $\VO^i(v_3,o_2)$ to 0. Once the $\VO^i$ matrix is updated for all the motivations $\times$ options inconsistencies, we compute the value ranking $R_{\mo}^i$. Algorithm~\ref{alg:MO} illustrates this procedure.

\begin{algorithm}[!htb]
\small
\SetAlgoLined
\KwIn{$M^i$, $\VO^i$, $C^i$, $V$}
\KwOut{$R_{\mo}^i$}
$\VO^i_{\mo} \leftarrow \VO^i$ \tcc*{Temporary copy, we need information from the original $\VO^i$ in the next loops}
\For{$m_{a} \in M^i : \; m_{a} \neq \varnothing$}{
    \For{$m_{b} \in M^i \setminus \{ m_{a} \}$}{
        $V_{\alpha} = V \setminus \{v : v \in m_{a}\} : \VO^i(v, o_a) == 1$    \tcc*{Values supporting $o_a$ in $\VO^i$, except values in $m_a$}
        \For{$v_x \in V_{\alpha}$ }{
            \uIf{$v_x \in m_{b}$}{
                \For{$v_y \in m_{a}$ }{
                    $V_{\beta} = V \setminus \{v : v \in m_{b}\} : \VO^i(v, o_b) == 1$ \tcc*{Values supporting $o_b$ in $\VO^i$, except values in $m_b$}
                    \uIf{$v_y \in V_{\beta}$}{
                        $\VO^i_{\mo}(v_x, o_a) = 0$;
                    }
                }
            }
        }
    }
}
$\VO^i \leftarrow \VO^i_{\mo}$ \;
$U^i = \VO^i \times C^{i^T}$\;
$R^i_{\mo} \, = \texttt{rank} (U^i)$\;
\caption{Method $MO$}
\label{alg:MO}
\end{algorithm}

\subsection{Disambiguation Strategy}
\label{sec:strategy-method}

The disambiguation strategy is intended to drive the interactions between AI agents and participants by addressing the detected inconsistencies between participants' choices and motivations, so as to improve the value preferences estimation process.
Inspired by popular AL strategies (Section~\ref{sec:active-learning-related-works}), the strategy iteratively targets the participants deemed to be most informative. We associate informativeness with the inconsistency between a participant's choices and motivations, assuming that the largest inconsistencies may reveal the biggest mistakes in the value preferences estimation process. By addressing the most informative participants first, we aim to rapidly improve the quality of value preferences estimation for all participants. \rebuttal{Such an approach can improve the estimation of individual value preferences---the more disambiguations are resolved, the more accurately value preferences are expected to be estimated. Furthermore, this approach would also positively impact the downstream application of aggregating value preferences at the population level by decreasing the total computations needed for a more accurate overall value preference estimation.}

Figure~\ref{fig:interaction-strategy} provides an overview of the proposed strategy. We consider a hybrid participatory setting where the AI agents are equipped with an NLP model tasked to predict the set of values mentioned in each participant's motivations. Then, value preferences are estimated on the basis of the participants' choices and the value labels that are predicted to support each motivation they provide.
We propose that AI agents iteratively interact with the participants with the largest detected inconsistencies between the value preferences estimated from their choices alone and the value preferences estimated from their motivations alone (provided in support of those choices).
In our method, the AI agents interact by asking whether the provided motivations have been correctly interpreted (i.e., if the predicted value labels are correct). Other interaction strategies can be implemented (e.g., querying the participants on whether the preference between two values $v_a$ and $v_b$ has been correctly estimated), which we discuss as future work (Section~\ref{sec:conclusions}).

\begin{figure}[ht]
    \centering
    \includegraphics[width=\linewidth]{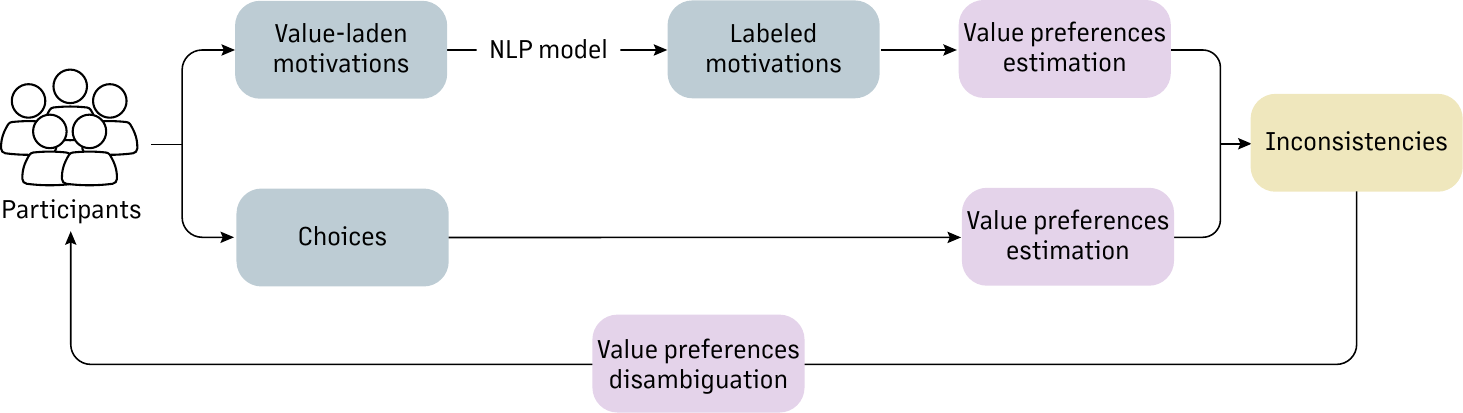}
    \caption{Overview of the proposed disambiguation strategy, guided by the detected inconsistencies between value preferences estimated from participants' choices and motivations.}
    \label{fig:interaction-strategy}
\end{figure}

Our setting is akin to an AL setting where value labels are iteratively retrieved \rebuttal{from an annotated dataset (in our experiments, the PVE dataset, as we further elaborate in Section~\ref{sec:AL-training-procedure})} to train a value classification NLP model. The most informative participants are iteratively selected by the strategy and asked to provide the correct value labels on their motivations, in practice treating the participants themselves as oracles.
At every iteration of the AL procedure, we use the current version of the NLP model to predict value labels on all the unlabeled motivations and use the predicted labels to estimate the value preferences of the participants whose motivations are not yet labeled, with both method $C$ and method $M$. Then, for each participant, we calculate the distance between the value ranking estimated with method $C$ and the value ranking estimated with method $M$. We use the Kemeny distance \shortcite{kemeny1962,Heiser2013} to measure the distance between rankings, as it accounts for potential ties between values (Def.~\ref{def:ranking}). The Kemeny distance ($d_K$) between two value rankings ($R^i_C, R^i_M$) is defined as:

\begin{equation*}
    d_K (R^i_C, R^i_M) = \frac{1}{2} \sum_{j=1}^n \sum_{k=1}^n |x_{(C)jk} - x_{(M)jk}|,
\end{equation*}

\noindent where $n$ is the number of objects (in our case, $n=5$ is the number of values), and $x_{(C)jk}$ is equal to $1$ if value $j$ is preferred over value $k$ in ranking $R^i_C$, equal to $-1$ in the reverse case, and equal to $0$ if the two values are equally preferred.
Finally, we choose as the next batch the $p$ participants with the largest Kemeny distance between the value rankings estimated with method $C$ and method $M$, and retrieve value labels for the motivations they provided. The NLP model is trained with the newly collected annotated motivations, and the AL strategy is re-iterated with the updated version of the NLP model.

\section{Experimental Setting}
\label{sec:experimental-setting}

We describe the experiments we perform to evaluate the proposed method\footnote{The code is available at \url{https://github.com/enricoliscio/value-preferences-estimation}}. First, we compare the five proposed value preferences estimation methods through a human evaluation procedure. Then, we use the best-performing value preferences estimation method in an AL setting to evaluate the disambiguation strategy by comparing it to traditional AL strategies.

\subsection{Value Preferences Estimation}
\label{sec:eval-value-preferences}

Given the participation process on energy transition using PVE described in Section \ref{sec:pve}, \rebuttal{we initialize $\VO^i$} by considering a value $v_l$ as relevant for an option $o_j$ if at least $t$ motivations (in our case, we set $t = 20$) among all participants were annotated with $v_l$ for $o_j$. \rebuttal{The resulting initial $\VO^i$ matrix (as shown in Table~\ref{tab:value-option-matrix}) is intended to represent an average rough estimate of the participants' value preferences. We use this as a starting point to apply the methods described in Section~\ref{sec:estimating-value-system} for all participants. In this way, we create a common starting point to show the effect that the different methods have on tailoring the resulting value rankings to each individual. Nevertheless, other choices for initializing $\VO^i$ are equally valid, as we elaborate in Section~\ref{sec:limitations}.}

\begin{table}[htb]
\centering
\small
\begin{tabular}{llllllll}
\centering
& & \multicolumn{6}{c}{\textbf{Options}} \\ 
& & $o_1$ & $o_2$ & $o_3$ & $o_4$ & $o_5$ & $o_6$ \\ \cline{3-8} 
\multirow{5}{*}{{\rotatebox[origin=c]{90}{\textbf{Values}}}} &
\multicolumn{1}{l|}{$v_1$} & \multicolumn{1}{l|}{1} & \multicolumn{1}{l|}{1} & \multicolumn{1}{l|}{1} & \multicolumn{1}{l|}{1} & \multicolumn{1}{l|}{1} & \multicolumn{1}{l|}{1} \\ \cline{3-8} 

& \multicolumn{1}{l|}{$v_2$} & \multicolumn{1}{l|}{1} & \multicolumn{1}{l|}{1} & \multicolumn{1}{l|}{0} & \multicolumn{1}{l|}{1} & \multicolumn{1}{l|}{1} & \multicolumn{1}{l|}{1} \\ \cline{3-8} 

& \multicolumn{1}{l|}{$v_3$} & \multicolumn{1}{l|}{1} & \multicolumn{1}{l|}{1} & \multicolumn{1}{l|}{1} & \multicolumn{1}{l|}{0} & \multicolumn{1}{l|}{0} & \multicolumn{1}{l|}{0} \\ \cline{3-8} 

& \multicolumn{1}{l|}{$v_4$} & \multicolumn{1}{l|}{1} & \multicolumn{1}{l|}{1} & \multicolumn{1}{l|}{1} & \multicolumn{1}{l|}{0} & \multicolumn{1}{l|}{0} & \multicolumn{1}{l|}{1} \\ \cline{3-8} 

& \multicolumn{1}{l|}{$v_5$} & \multicolumn{1}{l|}{1} & \multicolumn{1}{l|}{1} & \multicolumn{1}{l|}{0} & \multicolumn{1}{l|}{0} & \multicolumn{1}{l|}{1} & \multicolumn{1}{l|}{0} \\ \cline{3-8} 
\end{tabular}
\caption{\rebuttal{Initial} value-option matrix ($\VO^i$) for the energy transition PVE.}
\label{tab:value-option-matrix}
\end{table}

We analyze each method ($C$, $M$, $\tb$, $\mc$, and $\mo$) individually, and a sequential combination of the proposed methods in the following order: $\mo \Rightarrow \mc \Rightarrow \tb$. We choose this sequential combination for two reasons:
\begin{enumerate*}[label=(\arabic*)]
    \item the method $\tb$ should be executed last because it does not impact the $VO^i$ matrix directly and thus would not affect the subsequent methods, and
    \item \rebuttal{we tested both $\mo \Rightarrow \mc \Rightarrow \tb$ and $\mc \Rightarrow \mo \Rightarrow \tb$, and empirically found that the former consistently yields better results (see Appendix~A.2 for the comparison).}
\end{enumerate*}
To combine these methods sequentially, we use the ranking \rebuttal{and $\VO^i$} resulting from $\mo$ as input for $\mc$, and the ranking \rebuttal{and $\VO^i$} resulting from $\mc$ as input for $\tb$.
Finally, for the individual analysis of the methods $\tb$ and $\mc$, that require a previously estimated ranking, we start with the ranking estimated from choices alone (method $C$).
We evaluate these methods based on the resulting value preferences rankings, which we refer to as $R_C$, $R_M$, $R_{\tb}$, $R_{\mc}$, $R_{\mo}$, and \Rcomb (where \Rcomb is the result of the sequential combination $\mo \Rightarrow \mc \Rightarrow \tb$).

\subsubsection*{Evaluation Procedure}

Two evaluators, with previous knowledge of values and this specific PVE, were asked to independently judge the value preferences of a subset of participants based on their choices $C^i$ and the provided textual motivations (from which $M^i$ was annotated). We did not describe our value preference estimation methods to the evaluators.

The evaluators were presented with a \rebuttal{PVE} participant's choices and motivations \rebuttal{(but not to the value preferences estimated through our methods)}, proposed pairs of values (e.g., $v_a$ and $v_b$), and asked to \rebuttal{judge how the two values should be ranked for that participant} with the following options:
\begin{enumerate*}[label=(\arabic*)]
    \item $v_a \succ v_b$;
    \item $v_a \prec v_b$;
    \item $v_a \sim v_b$; or
    \item ``I do not know'', if they believe there is not enough information to make a proper comparison.
\end{enumerate*}
\rebuttal{Up to four different pairs of values $(v_a,v_b)$ were chosen for each selected PVE participant and judged by the evaluators,} with the intent of collecting sufficient information about a participant while increasing the number of analyzed participants.

The values to be compared were randomly selected from a set of value rankings that showed divergence across the methods. Our goal with this procedure is to assess the extent to which the proposed methods estimate value preferences similarly to the human evaluators. Within the envisioned application context described in Section~\ref{sec:introduction}, we expect that, as the methods' rankings mirror human intuition, they might provide meaningful feedback to participants in a participatory system.

\subsection{Disambiguation Strategy}
\label{sec:exp-setup-strategy}

We test the disambiguation strategy as a sampling strategy in an AL setting, where the motivations' annotations are iteratively retrieved and used to train an NLP model tasked to classify the values that support each motivation. We treat value classification as a multi-label classification task, where each motivation is annotated with zero or more value labels. Since not all provided motivations ought to be value-laden, a motivation may have zero labels in case none of the values in Table~\ref{tab:selected-values} is deemed relevant.

\subsubsection{Model Selection}

Multi-label BERT \shortcite{Devlin2019} has been shown to produce state-of-the-art performances on similar value classification tasks at the time of writing \shortcite{Liscio2022a,Kiesel2022,Huang2022,Qiu2022ValueNet:System}.
\rebuttal{As the PVE corpus was originally collected in Dutch, we chose to employ RobBERT \shortcite{delobelle-etal-2020-robbert}, a BERT variant considered state-of-the-art for the Dutch language at the time of writing.
However, due to the more widespread usage of the English language in NLP models, we also decided to translate the corpus to English and test two models trained in English---a RoBERTa model \shortcite{liu2019roberta} (similar to the Dutch model) and a comparably sized model with a different architecture, XLNet \shortcite{YangDYCSL19}. We further detail our experiments and hyperparameters search in Appendix~A.3, with Table~\ref{tab:F1} showing the performances with the best resulting models. As noticeable, the difference between the three tested models is minimal. Thus, we opted for the RobBERT Dutch model to employ the original data.}

\begin{table}[ht]
    \centering
    \small
    \begin{tabular}{lccc}
         \toprule
         & RobBERT (Dutch) & RoBERTa (English) & XLNet (English) \\
         \midrule
         micro $F_1$-score & 0.64 & 0.65 & 0.65 \\
         macro $F_1$-score & 0.63 & 0.64 & 0.64 \\
         \bottomrule
    \end{tabular}
    \caption{Micro and macro $F_1$-scores with the tested Dutch and English models.}
    \label{tab:F1}
\end{table}

\subsubsection{Training Procedure}
\label{sec:AL-training-procedure}

\rebuttal{In a typical AL setting, a large pool of unlabeled data points is initially available. A sampling strategy is used to iteratively select a set of unlabeled data points that are to be annotated and added to the pool of labeled data points (i.e., the data points that are used to train the NLP model). Together, labeled and unlabeled data points constitute the set of data points that are available for training. In addition, a set of labeled test data points (which are not available to be selected through the sampling strategy) is kept aside to evaluate the NLP model. In our case, we employ the annotated PVE dataset described in Section~\ref{sec:pve} to simulate the AL procedure. That is, we initially set aside a test set, and pretend that no labels are available for the remaining data points (which constitute the initial set of unlabeled data points). As the sampling strategy selects the unlabeled data points to be annotated, we retrieve the corresponding annotations from our PVE dataset and add these labeled data points to the set of labeled data points.}

At every AL iteration, we have a set of labeled motivations (whose labels have been retrieved and that are used to train the NLP model), a set of unlabeled motivations (whose labels can be retrieved if selected by the sampling strategy), and a set of test motivations (that are only used for evaluation). Analogously, \rebuttal{we refer to the PVE participants who wrote the motivations in the corresponding sets as} labeled participants, unlabeled participants, and test participants. At every iteration, the model is trained with the labeled motivations, and used to predict labels on the unlabeled motivations. With the predicted labels, the value preferences of the unlabeled participants are estimated. The disambiguation strategy is then used to select the $p$ unlabeled participants with the most inconsistent value preferences estimated from choices and from motivations alone. The $p$ participants are added to the set of labeled participants, the labels of the motivations provided by the participants are retrieved, and the motivations are added to the set of labeled motivations.

As is common in AL settings, we warm up the NLP model by initializing the set of labeled participants with 10\% of the available participants, and the set of labeled motivations with the motivations provided by those participants. At each iteration, we train the NLP model with the labeled motivations. We use the trained model to predict labels on the test motivations and use these labels to
\begin{enumerate*}[label=(\arabic*)]
    \item estimate the value preferences of the test participants with the best-performing value preferences estimation method, and
    \item evaluate the performance of the NLP model.
\end{enumerate*}
We then use the disambiguation strategy to select $p=39$ participants, so as to add 5\% of the available participants to the labeled participants set at each iteration. We iterate the procedure for 5 iteration steps and repeat it in a 10-fold cross-validation.

\subsubsection{Evaluation Procedure}

We evaluate how the proposed disambiguation strategy drives the NLP model performance and the estimation of value preferences, comparing it to the respective toplines and baselines.

We perform 10-fold cross-validation to measure the performance of the NLP model trained on all available data and use the result as the NLP topline during the AL procedure. We use a model trained on all data to predict labels on all the motivations and use the predicted labels to estimate all participants' value preferences with the best-performing value preferences estimation method. We treat the resulting value rankings as value preferences topline during the AL procedure, as they represent the best possible value rankings that can be estimated with the mistakes introduced by using the labels predicted by an NLP model instead of the ground truth annotations.
At every iteration of the AL procedure, we compare the NLP performance on the test set to the NLP topline, and the estimated value preferences of the test participants to the value preferences topline. For the NLP performance, we report \rebuttal{the micro $F_1$-score as it accounts for the label distribution (which is imbalanced, see Table~\ref{tab:annotation}).}
Finally, for the value preferences estimation performance, we report the Kemeny distance between the estimated value preferences of the test participants and the corresponding value preferences topline.

We compare the results to two baselines. First, we employ the uncertainty sampling strategy (Section~\ref{sec:active-learning-related-works}) to select 5\% of motivations (i.e., 145 motivations) at each iteration, similarly to the evaluated disambiguation strategy. This strategy is solely driven by motivations informativeness, ignoring the connection between the motivations and their authors. We choose this strategy as a baseline since traditional NLP AL strategies are solely driven by information about the NLP task, as described in Section~\ref{sec:active-learning-related-works}. 
Second, we employ a random baseline, where at each iteration 5\% random participants ($p=39$ participants, similarly to the proposed disambiguation strategy) and their motivations are added to the labeled set.
With both our proposed strategy and the baselines, we plot the trend of the NLP and value preferences estimation performances throughout the progressive iterations. We compare them with each other and to the corresponding toplines.

\section{Results and Discussion}
\label{sec:results}

We present and discuss the evaluation of the value preferences estimation methods and the interactive disambiguation strategy.

\subsection{Value Preferences Estimation}

When comparing the five proposed value preferences estimation methods, we aim to answer two questions:
\begin{enumerate*}[label=(\arabic*)]
    \item How well can each method estimate value preferences compared to humans?
    \item How does the estimation of value preferences differ among the methods proposed?
\end{enumerate*}

The evaluators performed 1047 comparisons. We discard the responses indicating that there was not enough information to judge values preference (``I do not know''), reducing the analyzed set to 766 total responses by either one of the evaluators. Figures~\ref{fig:evaluation_1} and \ref{fig:evaluation_2} present the performance of each method in terms of matching each evaluator's responses. 
These comparisons overlapped 269 times (i.e., the annotators performed the same comparisons). Considering this subset of overlapping comparisons, we find an agreement in 122 (45.35\%) and disagreement in 147 (54.65\%) comparisons, resulting in a Kappa score of 0.247, which is considered a fair agreement \shortcite{Landis1977TheData}.
To mitigate the effect of individual biases, in the remainder of the analysis, we focus on the pairwise comparisons that evaluators agreed on, as presented in Figure~\ref{fig:evaluation_both}.

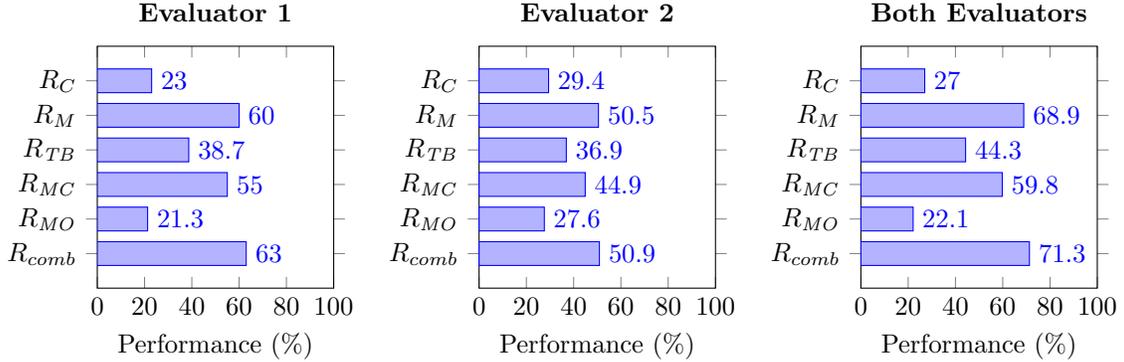
\begin{figure}[!htb]
\small
\centering
    \begin{subfigure}[b]{0.31\columnwidth}
        \begin{tikzpicture}
        \begin{axis}[
          width=\columnwidth,
          height=4.8cm,
          xbar,
          title={\textbf{Evaluator 1}},
            title style={align=center},
            ymin = 0, ymax = 7,
            ytick=data,
            yticklabels={\Rcomb, $R_{\mo}$, $R_{\mc}$, $R_{\tb}$, $R_M$, $R_C$},
            xmin = 0, xmax=100,
            xlabel = {Performance (\%)},
            nodes near coords,
            nodes near coords align={horizontal},
            bar width=9pt]
            \addplot
            coordinates { (63, 1) (21.3, 2) (55, 3) (38.7, 4) (60, 5) (23, 6)};
        \end{axis}
        \end{tikzpicture}
    \caption{Overlap with Evaluator 1.}
    \label{fig:evaluation_1}
    \end{subfigure}
    ~
    \begin{subfigure}[b]{0.31\columnwidth}
        \begin{tikzpicture}
        \begin{axis}[
          width=\columnwidth,
          height=4.8cm,
          xbar,
          title={\textbf{Evaluator 2}},
            title style={align=center},
            ymin = 0, ymax = 7,
            ytick=data,
            yticklabels={\Rcomb, $R_{\mo}$, $R_{\mc}$, $R_{\tb}$, $R_M$, $R_C$},
            xmin = 0, xmax=100,
            xlabel = {Performance (\%)},
            nodes near coords,
            nodes near coords align={horizontal},
            bar width=9pt]
            \addplot
            coordinates { (50.9, 1) (27.6, 2) (44.9, 3) (36.9, 4) (50.5, 5) (29.4, 6)};
        \end{axis}
        \end{tikzpicture}
    \caption{Overlap with Evaluator 2.}
    \label{fig:evaluation_2}
    \end{subfigure}
    ~
    \begin{subfigure}[b]{0.31\columnwidth}
    \begin{tikzpicture}
        \begin{axis}[
          width=\columnwidth,
          height=4.8cm,
          xbar,
          title={\textbf{Both Evaluators}},
            title style={align=center},
            ymin = 0, ymax = 7,
            ytick=data,
            yticklabels={\Rcomb, $R_{\mo}$, $R_{\mc}$, $R_{\tb}$, $R_M$, $R_C$},
            xmin = 0, xmax=100,
            xlabel = {Performance (\%)},
            nodes near coords,
            nodes near coords align={horizontal},
            bar width=9pt]
            \addplot
            coordinates { (71.3, 1) (22.1, 2) (59.8, 3) (44.3, 4) (68.9, 5) (27, 6)};
        \end{axis}
    \end{tikzpicture}
    \caption{Overlap where they agree.}
    \label{fig:evaluation_both}
    \end{subfigure}
\caption{Performance of the value preferences estimation methods, measured as the overlap with the evaluators' answers.}
\label{fig:evaluation}
\end{figure}

As Figure~\ref{fig:evaluation} displays, the rankings $R_{M}$, $R_{\mc}$, and \Rcomb provide the best performance in terms of human-like value preferences estimation. When compared to $R_C$, the combined method \Rcomb estimated value preferences 2.64 times more similarly to humans (considering the subset where evaluators agreed). Further, we observe that $R_{M}$ and $R_{\mc}$ also performs better than $R_C$. The only exception in terms of performance is $R_{\mo}$, which performs slightly worse than $R_C$. These findings show that combining choices and motivations in estimating value preferences can significantly increase the degree to which an automated method can estimate value preferences similarly to humans, with respect to using only choices.

Finally, we notice that the performance of $R_M$ is similar to the performance of \Rcomb. This is to be expected, as \Rcomb prioritizes motivations over choices, and $R_M$ only employs motivations to estimate value preferences. The visibly better performance of $R_M$ with respect to $R_C$ further motivates the need to consider textual motivations to estimate value preferences that are consistent with human evaluation. With our dataset, combining choices and motivations led to slightly better results than employing just the motivations. Further experiments with other data are needed to confirm this observation.

\subsubsection{Comparative Analysis}

For each method, we average the value preference rankings (that is, the position that the values have in the ranking that results after applying the method). We indicate with $\succ$ the values that have significantly different average rankings ($p \leq 0.05$) and with $\succeq$ the values that do not have significantly different averages. The following are the resulting average rankings per each different method:

\setlength{\columnsep}{-2.5cm}
\begin{multicols}{2}
\begin{itemize}
    \item $R_C$: $v_1 \succ  v_2 \succ v_4 \succ v_5 \succeq v_3$
    \item $R_M$: $v_3 \succ v_1 \succeq v_2 \succ v_5 \succeq v_4$
    \item $R_{TB}$: $v_1 \succ v_2 \succ v_4 \succ v_5 \succ v_3$
    \item $R_{MC}$: $v_1 \succ v_2 \succ v_5 \succ v_3 \succ v_4$
    \item $R_{MO}$: $v_1 \succ v_2 \succ v_4 \succ v_5 \succeq v_3$
    \item \Rcomb: $v_1 \succeq v_2 \succ v_5 \succeq v_3 \succeq v_4$
\end{itemize}
\end{multicols}

Method $C$ ranked the value $v_1$ as the most important for all individuals, regardless of their choices, due to the characteristics of the initial value option-matrix ($\VO^i$) in Table~\ref{tab:value-option-matrix}, which considers $v_1$ relevant for all choice options. As we attribute the minimum ordinal ranking for the values in case of ties (Def.~\ref{def:ranking}), any choices would lead to $R^i_C$ with $v_1$ as (one of) the most important value(s), except for method $M$ which does not consider choices.

Let $R_C$ be a baseline for comparison. Figure~\ref{fig:changes} indicates how many positions the final ranking changed across values (we do not consider method $M$ since it did not use $R_C$ as baseline). For example, consider two rankings $R_1: v_1 \succ v_2 \succ v_3 \succ v_4 \succ v_5$ and $R_2: v_2 \succ v_3 \succ v_1 \succ v_4 \succ v_5$. We consider four position changes from $R_1$ to $R_2$: $v_1$ changed from the first to the third position (two changes), $v_2$ changed from the second to the first position (one change), and $v_3$ changed from the third to the second position (one change).

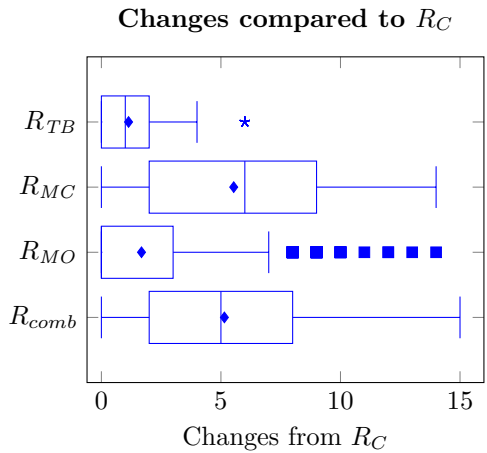
\begin{figure}[!htb]
\centering
\small
    \begin{tikzpicture}
    \begin{axis}[
    width=0.45\columnwidth,
    boxplot/draw direction=x,
    title={\textbf{Changes compared to $R_C$}},
	ytick={1,2,3,4},
	yticklabels={\Rcomb, $R_{\mo}$, $R_{\mc}$, $R_{\tb}$},
    ymin = 0, ymax = 5,
    xmin=-0.6, xmax=16,
	xlabel={Changes from $R_C$},
	xlabel style={align=center},
	boxplot/average=auto,
	]
 	\addplot+[blue,boxplot,mark options={fill=blue}]
 	table[y={ER_C_MO_MC_TB}] {tikz/changes.dat};
 	\addplot+[blue,boxplot,mark options={fill=blue}]
    table[y={ER_C_MO}] {tikz/changes.dat};
 	\addplot+[blue,boxplot,mark options={fill=blue}]
    table[y={ER_C_MC}] {tikz/changes.dat};
    \addplot+[blue,boxplot,mark options={fill=blue}]
    table[y={ER_C_TB}] {tikz/changes.dat};
    \end{axis}
\end{tikzpicture}
\caption{Average changes in the value rankings when compared to $R_C$.}
\label{fig:changes}
\end{figure}

Rankings $R_{\tb}$ and $R_{\mo}$ barely deviate from the average $R_{C}$. Instead, $R_\mc$ and the combined approach \Rcomb show significant deviation from $R_C$, indicating a larger difference at an individual value preferences level.
The large deviation and the good performance (see Figure~\ref{fig:evaluation}) of these two methods suggest that they estimate individually tailored value preferences that align with human intuition.

\subsection{Disambiguation Strategy}

First, we report the results of the toplines. The NLP topline resulted in an average micro $F_1$-score of 0.64 (Table~\ref{tab:F1}), which is slightly lower than similar value classification tasks \cite{Liscio2022a,Huang2022}, likely due to the smaller dataset size. For the value preferences topline, we use the predicted motivation labels to estimate value rankings through the \Rcomb method (the best-performing value preferences estimation method). The value preferences topline resulted in an average Kemeny distance of 1.88 (with 2.88 standard deviation) from the value rankings estimated with the $\mo \Rightarrow \mc \Rightarrow \tb$ method (with the resulting ranking \Rcomb) by using the ground truth annotations on the motivations. We use these toplines to measure the trend of the results throughout the AL iterations.

We report the results of our experiments in Figures~\ref{fig:NLP-performance} and \ref{fig:value-estimation-performance}. In all experiments, at every iteration we used the tested strategy to select 5\% of the data to be added to the set of labeled data. However, since different participants provided different numbers of motivations, selecting the motivations provided by 5\% of the participants may not correspond to 5\% of all available motivations. In Figures~\ref{fig:NLP-performance} and \ref{fig:value-estimation-performance}, we show on the x-axis the number of motivations used for training the NLP model at the corresponding iteration. While that corresponds to exactly 5\% increments in the case of the uncertainty strategy (which selects 5\% of the motivations at every iteration), it is not the case for the random and disambiguation strategies (which selects 5\% of the participants at every iteration).

\begin{figure}[htb]
    \centering
    \small
    \begin{minipage}{.48\textwidth}
        \centering
\begin{tikzpicture}
    \begin{axis}[
        width=0.9\linewidth,
        title={\textbf{NLP Classification}},
        xlabel = {\% of data used for training},
        ylabel = {micro $F_1$-score},
        ymin = -0.05, ymax = 0.68,
        minor y tick num = 1,
       xtick={10,15,20,25,30,35},
        xmin = 8, xmax = 38,
        xticklabels={10\%, 15\%, 20\%, 25\%, 30\%, 35\%},
        legend pos=south east,
        legend cell align={left},
    ]
    \addplot+ [
    error bars/.cd,
    x dir=both, x explicit,
    y dir=both, y explicit,
    ] table[y error=std] {tikz/f1_score/random.dat};
    \addlegendentry{random}
    \addplot+ [
    error bars/.cd,
    x dir=both, x explicit,
    y dir=both, y explicit,
    ] table[y error=std] {tikz/f1_score/uncertainty.dat};
    \addlegendentry{uncertainty}
    \addplot+ [
    every mark/.append style={solid, fill=teal},
    mark=triangle*][
    teal,
    error bars/.cd,
    x dir=both, x explicit,
    y dir=both, y explicit,
    ] table[y error=std] {tikz/f1_score/disambiguation.dat};
    \addlegendentry{disambiguation}
    \addplot+[dashed,magenta,mark=none]
    coordinates {(-10,0.64) (100,0.64)};
    \addlegendentry{topline}
    \end{axis}
\end{tikzpicture}
        \caption{NLP performance (micro $F_1$-score), compared to the NLP topline (dashed horizontal line).}
        \label{fig:NLP-performance}
    \end{minipage}
~~~
\begin{minipage}{.48\textwidth}
    \centering
    \begin{tikzpicture}
    \begin{axis}[
        width=0.9\linewidth,
        title={\textbf{Value Preferences Estimation}},
        xlabel = {\% of data used for training},
        ylabel = {Distance from topline},
        ymin = 2.5, ymax = 9.5,
        minor y tick num = 1,
        xtick={10,15,20,25,30,35},
        xmin = 8, xmax = 38,
        xticklabels={10\%, 15\%, 20\%, 25\%, 30\%, 35\%},
        legend pos=north east,
        legend cell align={left},
    ]
    \addplot+ [
    error bars/.cd,
    x dir=both, x explicit,
    y dir=both, y explicit,
    ] table[y error=std] {tikz/profiles/random.dat};
    \addlegendentry{random}
    \addplot+ [
    error bars/.cd,
    x dir=both, x explicit,
    y dir=both, y explicit,
    ] table[y error=std] {tikz/profiles/uncertainty.dat};
    \addlegendentry{uncertainty}
    \addplot+ [
    every mark/.append style={solid, fill=teal},
    mark=triangle*][
    teal,
    error bars/.cd,
    x dir=both, x explicit,
    y dir=both, y explicit,
    ] table[y error=std] {tikz/profiles/disambiguation.dat};
    \addlegendentry{disambiguation}
    \end{axis}
\end{tikzpicture}
\caption{Value preferences estimation performance, measured as average Kemeny distance from the value preferences topline.}
\label{fig:value-estimation-performance}
\end{minipage}
\end{figure}

The random strategy has a varying step size that roughly averages to 5\%, as expected by a strategy that randomly selects participants. Instead, the step size of the disambiguation strategy is consistently smaller than the other two (for this strategy we plot six steps, as opposed to five for the other strategies), meaning that at every iteration the strategy chooses participants who have provided less motivations than the average participant. This empirically matches the intuition behind the strategy---participants who have provided few motivations have a $R_M$ (value ranking calculated from motivations alone) that is mostly composed of ties between values. Such undetermined $R_M$ have a large distance from the corresponding $R_C$, which instead considers all the choices provided by the participants.

The NLP performances of the model trained with the disambiguation strategy and with the two baseline strategies (uncertainty and random) are illustrated in Figure~\ref{fig:NLP-performance}. No significant difference between the compared methods is visible, as all three strategies lead to a rapid improvement in performances that approaches the NLP topline when roughly 30\% of the available motivations are used for training.
In line with these results, experimental findings \shortcite{ein-dor-etal-2020-active} show that there is no single AL strategy that outperforms all others across different datasets, and, in some cases, no significant difference is observable with the random strategy.
Ultimately, these results demonstrate that the proposed disambiguation strategy, despite being guided by the downstream task of estimating value preferences, does not significantly affect the NLP performance.

Figure~\ref{fig:value-estimation-performance} presents the value preferences estimation performance of the three compared strategies, measured as the average Kemeny distance between the value preferences estimated with the labels predicted by the current iteration of the NLP model and the value preferences topline.
First, we remind that the topline has been calculated with the label predictions resulting from a model trained with all available data. However, the training process with the tested strategies is performed as 10-fold validation, thus a different subset of the dataset is used for training in each fold. Consequently, we do not expect the Kemeny distance to approach zero, as different data was used during the training process (thus resulting in different individual value preferences). Still, the topline reference allows us to compare the value preferences estimation performance trend of the three strategies.

We observe that the value preferences estimation performance trend is similar for all three strategies, leading to a rapid decrease in distance from the topline that mirrors the rapid improvement in the $F_1$-score. 
While the results are comparable when 20\% or more motivations are used for training, the results with 
$\sim$15\% of the training data show small differences---while the $F_1$-score performances at this stage are almost identical, there is a small difference in value preferences estimations. In particular, the uncertainty strategy (which ignores the link between users and motivations) is worse than the other two tested strategies, which motivates the usage of a user-driven strategy instead of a motivation-driven strategy. However, the differences are not sufficiently large to draw a definitive conclusion.

Overall, we notice no significant difference between the proposed strategy and the baselines. We discuss two possible reasons.
First, the NLP performance is the biggest driver of value preferences estimation performance---in practice, the more motivations are correctly labeled, the more accurate the value preferences estimation is. With the analyzed data and the relatively small dimension of the dataset, no significant difference is noticeable between the tested strategies in NLP performance, including the random strategy, resulting in a similar trend in the value preferences estimation performance.
Second, the distance between $R_M$ and $R_C$ may not be the best indicator for the informativeness of a user. Considering the annotations from Section~\ref{sec:pve}, there is a distance of 8.0 (with 3.5 standard deviation) between $R_M$ and $R_C$ estimated for the same users. Thus, large distances between the two rankings may not be particularly informative in this dataset.
However, we believe that a strategy driven by the downstream application may be particularly useful in similar settings, as we elaborate as Future Work.

\section{Limitations}
\label{sec:limitations}

\rebuttal{We discuss the main limitations of our experimental results.}

\rebuttal{First, we discuss the generalizability of our experimental setting. As described in Section~\ref{sec:introduction}, we envision our value estimation and disambiguation methods to be employed during an ongoing deliberation. 
However, to showcase the proposed methods, we tested them on a concluded survey and used third-party annotations to evaluate them, thus limiting the validity of our results due to the subjective nature of the annotation and evaluation task. Such limitations can be addressed, for example, by asking annotators to perform vicarious annotations \shortcite{weerasooriya-etal-2023-vicarious} or by turning to a diverse sample of annotators \shortcite{ACAL}. However, the most effective approach would be to consult the participants themselves, as we further elaborate in Section~\ref{sec:conclusions}.} 

\rebuttal{Second, we discuss the initialization of $\VO^i$. We decided to initialize all initial $\VO^i$'s identically (as motivated in Section~\ref{sec:eval-value-preferences}) to demonstrate the effectiveness of the value preferences estimation strategies in tailoring value preferences to the individuals. However, in different settings, other initialization strategies could be equally valid. For instance, $\VO^i$ could be initialized based on 
\begin{enumerate*}[label=(\arabic*)]
    \item the results of a previous (or another) session of deliberation,
    \item a self-reported estimate of value preferences from the participants,
    \item an initial estimate from the policy-makers, or
    \item a demographic grouping of initially estimated preferences.
\end{enumerate*}
The effectiveness of our proposed value preferences estimation methods ought to be studied with different initializations of $\VO^i$. Methods $\mo$ and $\mc$ adjust $\VO^i$ by turning ones into zeros---precisely, the best-performing method $comb$ leads the average $\VO^i$ matrix from having $21$ to having $15.71 \pm 3.74$ ones. Despite resulting in still populated $\VO^i$ matrices in our experiments, this effect could be detrimental with a less populated initial $\VO^i$. Similarly, different strategies (or applications thereof) should be devised for repeated use over several deliberation sessions, both to avoid matrix sparsity and that the latest rounds of deliberation (accidentally) override the results obtained in the previous.}

\rebuttal{Third, we discuss the choice of the value labels that we employ in our experiments. As described in Section~\ref{sec:pve}, we perform our experiments with a subselection of the values identified by \shortciteauthor{kaptein2020participatory}~\citeyear{kaptein2020participatory}. We choose so as our experiments are not intended as a comprehensive analysis of the value preferences of the survey participants, but rather a simplified scenario to showcase our methods. The application of our methods in a real deliberation setting ought to be able to handle \begin{enumerate*}[label=(\arabic*)]
    \item larger lists of values (to which our methods are compatible), and
    \item changing lists of values, which may be iteratively updated during the deliberation with methods such as the one proposed by \shortciteauthor{Liscio2022axiesJAAMAS}~\citeyear{Liscio2022axiesJAAMAS}.
\end{enumerate*}
Furthermore, in the dataset we used, values were annotated only when supporting the motivations and thus the related choice. However, choices could also demote values, and as so be reflected in the related motivations. Recent works have investigated this approach to value valence (i.e., that values could be promoted or demoted by actions) for value preferences aggregation \shortcite{lera2024aggregating} and value classification in text \shortcite{Sorensen2024}. In this case, our methods ought to be updated. First, the NLP model should be trained to predict a value label behind the motivations that range from positive to negative. Next, the notion of valence could be inserted into the disambiguation strategy (e.g., by prioritizing participants with motivations with opposite valences for the same values) and preferences estimation strategy (e.g., by accounting for the valence of the values when addressing inconsistencies).}

\rebuttal{Finally, we discuss the validity of our machine learning experiments.
We experimented on a (relatively small) dataset composed of survey answers in Dutch. Further experiments are needed to validate our findings with other types of data (e.g., conversational) and under-represented languages, and to validate the impact of label distribution on both the results with the NLP model and disambiguation strategy.
Furthermore, the proposed disambiguation strategy is sensitive to outlier participants, since it targets the largest inconsistencies between participants' choices and motivations. This creates the risk that the NLP model---and, consequently, the value preferences estimation results---are built on data that is not representative of the overall population, or worse, on noisy data. While the distinction between noisy and minority voices in subjective tasks is under debate \shortcite{plank-2022-problem,Cabitza_Campagner_Basile_2023}, our envisioned application addresses the problem by consulting the participants themselves, as we further elaborate in Section~\ref{sec:conclusions}.}

\section{Conclusion and Future Directions}
\label{sec:conclusions}

We introduce a method for estimating how participants prioritize competing values in a hybrid participatory system, through a disambiguation strategy aimed at guiding the interactions between AI agents and participants. Our method directly targets the detected inconsistencies between participants' choices and motivations.

First, we propose and compare methods for an AI agent to estimate the value preferences of individuals from one's choices and value-laden motivations, with the goal of generating an ordered value ranking within the analyzed context.
We aim to improve the estimation of value preferences by prioritizing value preferences estimated from motivations over value preferences estimated from choices alone. We test our methods in the context of a large-scale survey on energy transition. Through a human evaluation, we show that incorporating motivations to deal with conflicts in value preferences improves the performance of value preferences estimation by more than two times (in terms of similarity to human evaluators' value preferences estimation) and yields preferences that are more individually tailored.

Second, we propose a disambiguation strategy to drive the interactions between AI agents and participants, with the intent of improving the value preferences estimation performance. Our strategy prioritizes the interaction with the participants whose value preferences estimated from choices alone are most different from the value preferences estimated from motivations alone, following the rationale that such participants would be the most informative for rapidly adjusting and improving the value preferences estimation process. However, our results show no significant difference with compared baseline strategies, including a strategy where interactions with users are randomly determined.

Despite the inconclusive results, we believe that our proposed disambiguation strategy opens novel research avenues. Such a hybrid approach to an interaction strategy for value preferences disambiguation can help \rebuttal{not only} iteratively address algorithmic mistakes, \rebuttal{but also} foster self-reflection in participants \rebuttal{by situating their estimated value preferences in specific contexts and choices \shortcite{Liscio2023}, which has been shown to raise awareness and lead to changing perspectives} \shortcite{Lim2019FacilitatingConditions}. A strategy driven by the downstream task of value preferences estimation helps in integrating the different components involved in the value preferences estimation process (value label classification and aggregation of one's choices and motivations). 
Further, different disambiguation approaches could be tested. For instance, the strategy could target the participants with the most different choice distribution from the average, or with the largest amount of ties in their estimated value rankings.

We identify additional directions for future work. On the one hand, we suggest exploring other approaches to associate values with choice options beyond a binary matrix\rebuttal{---for instance, given $n$ considered values, by using an $n \times n$ matrix that reflects pairwise comparisons among all values, thus allowing non-transitive value preferences \shortcite{Alos-Ferrer2022Identifying}.} 
On the other hand, \rebuttal{using our proposed methods during an ongoing deliberation would open additional future work avenues. In such a setting, we envision a language model to be trained to recognize values in text through the disambiguation strategy, use it to classify the values in the motivations, and use this information to estimate value preferences.}
\rebuttal{An interesting extension compatible with the proposed methods is to let} participants themselves provide direct feedback to the AI agent, instead of relying on external evaluators. Additionally, \rebuttal{following the self-reflection fostered by the disambiguation strategy,} participants may be offered the option to adjust \rebuttal{their choices or} the estimated value preferences directly, instead of being limited to providing the correct value label supporting their motivations. Machine learning methods could then be employed for value preferences estimation, learning directly from the feedback provided by the participants.

Our work has the potential to contribute to value alignment between AI and humans. The estimated value preferences can serve as a starting point for the operationalization of values, e.g., for the synthesis of value-aligned normative systems \shortcite{Serramia2021,Montes2022}, as a foundation for international regulatory systems \shortcite{Bajgar2023}, or to formulate ethical principles through a combination of machine learning and logic \shortcite{Kim2021}. \rebuttal{In the context of a hybrid participatory system,} the estimated individual value preferences can be aggregated at a societal level \shortcite{lera2024aggregating} to provide policy-makers with an overview of the value preferences of a population.

\section*{Acknowledgments}

Enrico Liscio and Luciano C. Siebert contributed equally to this work.
This work was partially supported by TU Delft's AiTech initiative, by TAILOR, a project funded by the EU Horizon 2020 research and innovation programme under GA No 952215, and by the Netherlands Organisation for Scientific Research (NWO) through the Hybrid Intelligence Centre via the Zwaartekracht grant (024.004.022).
We thank Lionel Kaptein, Shannon Spruit, and Jeroen van den Hoven for their help with previous iterations of this work.

\vskip 0.2in
\bibliographystyle{theapa}

 \providecommand{\url}[1]{#1}

\end{document}